\documentclass[10pt,twocolumn,letterpaper]{article}

\usepackage[pagenumbers]{cvpr} 
\usepackage{colortbl}
\usepackage{color}
\definecolor{Gray}{gray}{0.9}
\usepackage{mdwmath}
\usepackage{eqparbox}
\usepackage{epsfig}
\usepackage{graphicx}
\usepackage{amsmath}
\usepackage{amssymb}
\usepackage{pifont}
\usepackage{url}            
\usepackage{booktabs}       
\usepackage{tabu}           
\usepackage{multirow}
\usepackage{graphicx}
\usepackage{algorithm}
\usepackage{algorithmicx}
\usepackage{algpseudocode}
\usepackage{amsmath,amssymb}
\usepackage{array}

\usepackage{cite}

\usepackage{makecell}
\usepackage{tabularx}
\usepackage{pifont}
\usepackage{multicol}
\usepackage{array}
\usepackage{adjustbox}

\usepackage{url}            
\usepackage{booktabs}       
\usepackage{amsfonts}       
\usepackage{nicefrac}       
\usepackage{microtype}      
\usepackage{graphicx}
\usepackage{amsmath}
\usepackage{bm}
\usepackage{multirow}
\usepackage{tabu}
\usepackage{siunitx}
\usepackage{adjustbox}
\usepackage{subcaption}
\usepackage{siunitx}
\usepackage{colortbl}
\usepackage{color}
\usepackage{pifont}
\definecolor{Gray}{gray}{0.9}
\definecolor{Lightorange}{RGB}{255,214,169}
\definecolor{Cyan}{rgb}{0.88,1,1}
\definecolor{reminder}{RGB}{255,0,0}
\definecolor{Red}{RGB}{176,36,24}

\usepackage{tabu}           
\usepackage{multirow}
\usepackage{booktabs}
\usepackage{makecell}
\usepackage{pifont}

\newcolumntype{x}[1]{>{\centering\arraybackslashå}p{#1pt}}
\newlength\savewidth

\newcommand{\PreserveBackslash}[1]{\let\temp=\\#1\let\\=\temp}
\newcolumntype{C}[1]{>{\PreserveBackslash\centering}p{#1}}
\newcolumntype{L}[1]{>{\PreserveBackslash\raggedright}p{#1}}

\definecolor{cvprblue}{rgb}{0.21,0.49,0.74}
\usepackage[pagebackref,breaklinks,colorlinks,allcolors=cvprblue]{hyperref}

\def\paperID{xxx} 
\def\confName{CVPR}
\def\confYear{2025}

\title{Insight-V: Exploring Long-Chain Visual Reasoning with \\ Multimodal Large Language Models}

\author{Yuhao Dong\textsuperscript{1}\thanks{Authors contributed equally to this research.~~\textsuperscript{\dag}Corresponding authors.},~~Zuyan Liu\textsuperscript{2,3}\footnotemark[1],~~Hai-Long Sun\textsuperscript{2,4},~~Jingkang Yang\textsuperscript{1},\\Winston Hu\textsuperscript{2},~~Yongming Rao\textsuperscript{2,3}$^{\dagger}$,~~Ziwei Liu\textsuperscript{1}$^{\dagger}$ \\
\textsuperscript{1}~S-Lab, NTU~~\textsuperscript{2}~Tencent~~\textsuperscript{3}~Tsinghua University~~\textsuperscript{4}~Nanjing University\\ \\
\textbf{\url{https://github.com/dongyh20/Insight-V}}
}
\begin{document}
\maketitle
\begin{abstract}

Large Language Models (LLMs) demonstrate enhanced capabilities and reliability by reasoning more, evolving from Chain-of-Thought prompting to product-level solutions like OpenAI o1.  Despite various efforts to improve LLM reasoning, high-quality long-chain reasoning data and optimized training pipelines still remain inadequately explored in vision-language tasks. In this paper, we present \textbf{Insight-V}, an early effort to \textbf{1)} scalably produce long and robust reasoning data for complex multi-modal tasks, and \textbf{2)} an effective training pipeline to enhance the reasoning capabilities of multi-modal large language models (MLLMs). Specifically, to create long and structured reasoning data without human labor, we design a two-step pipeline with a progressive strategy to generate sufficiently long and diverse reasoning paths and a multi-granularity assessment method to ensure data quality. 
We observe that directly supervising MLLMs with such long and complex reasoning data will not yield ideal reasoning ability. To tackle this problem, we design a multi-agent system consisting of a reasoning agent dedicated to performing long-chain reasoning and a summary agent trained to judge and summarize reasoning results. We further incorporate an iterative DPO algorithm to enhance the reasoning agent's generation stability and quality. 
Based on the popular LLaVA-NeXT model and our stronger base MLLM, we demonstrate significant performance gains across challenging multi-modal benchmarks requiring visual reasoning. Benefiting from our multi-agent system, Insight-V can also easily maintain or improve performance on perception-focused multi-modal tasks. 

\end{abstract}    
\section{Introduction}
\label{sec:intro}

\begin{figure}[t]
\centering
\includegraphics[width=0.5\textwidth]{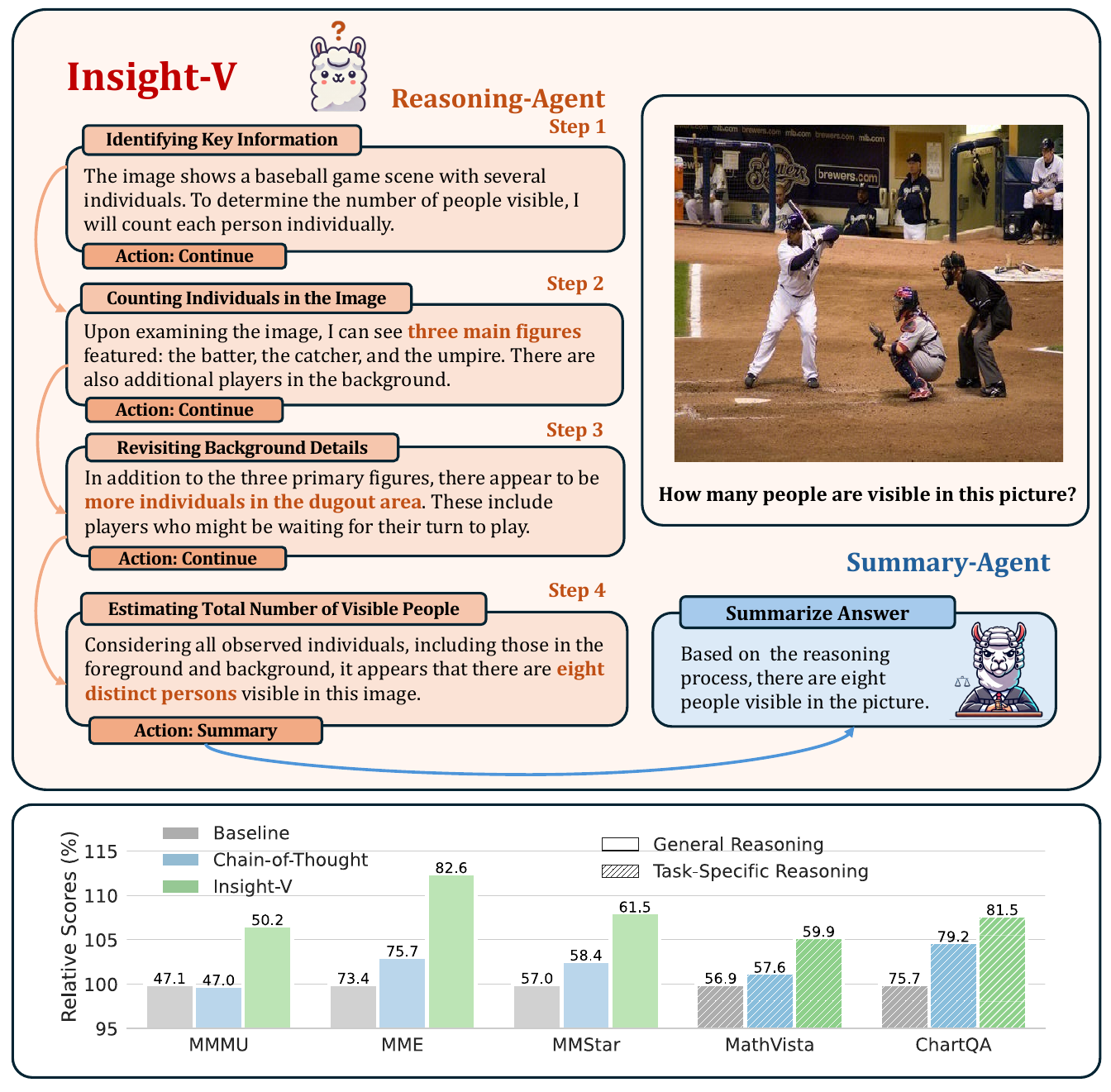} 
\caption{\textbf{Illustration and Performance of Insight-V.} Insight-V consists of two agents, one dedicated to reasoning and the other to summarization, driving significant improvements in performance across various visual reasoning benchmarks.}
\label{fig:teaser}
\vspace{-10pt}
\end{figure}

The development of artificial general intelligence requires models that can seamlessly understand and respond to multi-modal data. Recent advancements in Large Language Models (LLMs)~\citep{GPT4o,qwen2,dubey2024llama,qwen2.5} and Multi-modal LLMs (MLLMs)~\citep{liu2024llava,liu2024llava15,liu2024llavanext,chen2024internvl,qwen2vl,lu2024deepseek,yao2024minicpmv} have significantly facilitated this progress across various fields, ranging from common question-answering~\citep{qwen2vl,chen2024internvl,liu2024llavanext,li2024llavaov,liu2024oryx} to autonomous driving~\citep{tian2024drivevlm,ma2023dolphins} and robotics~\citep{yang2023octopus,driess2023palm}. Despite the substantial progress made in enhancing the performance of MLLMs on a wide range of tasks, enabling MLLMs to perform human-level reasoning remains a key challenge. This area remains underexplored and has yet to fully realize its potential.

Existing efforts~\citep{wei2022chain,yao2024tree} to enhance the reasoning capabilities of LLMs through long-chain reasoning have demonstrated considerable progress, largely benefiting from the availability of structured, high-quality data and well-established training pipelines. In contrast, teaching MLLMs to perform long-chain visual reasoning remains a significant challenge, primarily due to the lack of large-scale, high-quality datasets and efficient and effective training strategies. Compared to text-only data, visual reasoning data is not only more expensive to collect but also requires significant human labor for detailed annotation and validation, due to the absence of an effective data generation pipeline. Moreover, while previous work~\citep{zhang2023multimodal} has demonstrated that directly applying chain-of-thought~\citep{wei2022chain} reasoning can improve the capabilities of MLLMs, other research~\citep{zhang2024mavis,zhang2024improve} suggests that current training approaches have limited effectiveness in enhancing CoT reasoning. This highlights the inability of current MLLMs to leverage visual cues for precise step-by-step problem-solving, emphasizing the need for an effective training procedure that enables MLLMs to reason in detail while maintaining clear visual perception.

To address these challenges, we propose Insight-V, which incorporates two innovative designs to enhance reasoning capabilities. First, we introduce a data generation pipeline consisting of two key steps: a progressive strategy to generate structured, long-chain reasoning data with diverse reasoning paths, and a multi-granularity assessment system to evaluate and score these paths at different levels. Through automatic generation, assessment, and ranking strategies, the pipeline effectively operates without the need for human labor and makes the reasoning dataset more scalable for enhancing reasoning capabilities. To further improve MLLM reasoning beyond data scaling, we design a multi-agent system, as illustrated in Figure~\ref{fig:teaser}, that decomposes the problem-solving process into two distinct steps: reasoning and summarization. The reasoning agent generates a detailed reasoning process for the input query, while the summarization agent identifies key information within the reasoning process and selectively answers the question. To refine the quality of the reasoning, we employ an iterative DPO approach to enhance reasoning capabilities. The two agents collaborate to further improve the reasoning quality. Our findings demonstrate that this system significantly enhances the performance of various MLLMs across a broad range of visual reasoning benchmarks.

We evaluate Insight-V by integrating our system into the widely used LLaVA-NeXT~\citep{liu2024llavanext} model. To further highlight the potential of our design in advancing state-of-the-art models, we construct a robust base MLLM and apply our system to demonstrate its ability to enhance SOTA models across various visual reasoning benchmarks. Notably, Insight-V improves LLaVA-NeXT in an average performance of 7.0\% across seven challenging visual reasoning benchmarks, while a 2.9\% improvement is observed with the strong base MLLM, underscoring the effectiveness and generalizability of Insight-V.

In summary, Insight-V offers \textbf{1)} a scalable data generation pipeline for long-chain, high-quality reasoning data, \textbf{2)} a multi-agent system that decomposes visual reasoning tasks into reasoning and summarization, and \textbf{3)} a two-stage training pipeline to enhance visual reasoning capabilities. Together, these contributions address key challenges in visual reasoning, providing a solid foundation for future research in MLLM reasoning.
\section{Related Work}
\label{sec:related}

\subsection{Vision-Language Reasoning}
Recent advancements in MLLMs~\citep{liu2024llava,liu2024llava15,liu2024llavanext,lin2023vila,bai2023qwenvl,lu2024deepseek,qwen2vl,liu2024oryx,li2024llavaov} have equipped these models with robust capabilities across diverse domains, including visual understanding~\citep{lin2023vila, qwen2vl}, mathematics~\citep{liang2023unimath}, college-level questions~\citep{chen2024internvl}, and scientific inquiries. In visual understanding, most research~\citep{liu2024llavanext,li2023monkey,tong2024cambrian,xu2024llavauhd,liu2024chain} emphasizes fine-grained detail analysis and localization, training models to perform visual reasoning with tailored datasets to enhance interpretive capabilities. For mathematics and expert-level reasoning, existing methods~\citep{gao2023g,zhang2024mavis,zhang2024improve} predominantly derive from Chain-of-Thought~\citep{wei2022chain} approaches, training MLLMs to generate step-by-step reasoning across various subjects. However, these approaches often focus primarily on improving dataset quality through Chain-of-Thought, overlooking the importance of structured reasoning paths and extended reasoning chains in advancing model reasoning capabilities. Additionally, significant challenges arise when relying on a single model to manage the entire reasoning process for complex tasks, underscoring the need for a multi-agent system to decompose and enhance this process. In this work, we tackle these challenges by introducing a scalable reasoning data generation pipeline and implementing a multi-agent system for reasoning and summarization decomposition, enhancing the overall reasoning capabilities of existing MLLMs.

\subsection{Vision-Language Alignment}
To align the model more closely with human preferences, several alignment techniques are employed for MLLMs. A widely used approach is Reinforcement Learning from Human Feedback~\citep{bai2022training} (RLHF), which iteratively refines the model’s responses based on human feedback, enhancing both response quality and interpretability. To further improve MLLM capabilities, Direct Preference Optimization~\citep{rafailov2024direct} (DPO) is introduced to simplify the alignment process. By directly training on human preference data, DPO optimizes the model's outputs to better match human-selected responses. However, traditional DPO is primarily focused on offline scenarios, and as the model evolves, the effectiveness of this approach may significantly diminish. To address this, Iterative DPO~\citep{chen2024self} has been proposed, which optimizes preference pairs through DPO at each iteration. It then generates new preference pairs for the next iteration using the updated model and evaluates them with a reward model. In this paper, we use iterative DPO to achieve stronger alignment and enhance the model's reasoning capabilities.
\section{Methodology}
\label{sec:method}
In this section, we provide a comprehensive description of the proposed Insight-V system, detailing its architecture and key contributions. Section~\ref{sec:overview} presents an overview of Insight-V, highlighting the core concepts of our approach. The design of Insight-V is structured around three primary components: \textbf{1)} a carefully constructed pipeline for structured, scalable reasoning data generation, as described in Section~\ref{sec:data}; \textbf{2)} a multi-agent MLLM system that supports complex visual reasoning, as detailed in Sections~\ref{sec:model}; and \textbf{3)} a streamlined yet effective training pipeline to enhance overall performance, as outlined in Section~\ref{sec:implementation}. Together, these components form a cohesive system that effectively tackles the challenges of performing detailed, long-chain reasoning while preserving visual perception capabilities.

\subsection{Overview}
\label{sec:overview}
Advancing reasoning capabilities of LLMs has been a focal point of extensive research. Despite these efforts, the reasoning potential within multi-modal LLMs has barely been explored. Most approaches aim to strengthen reasoning at the inference stage, assuming the model has already acquired robust reasoning skills. Other approaches optimize model parameters using chain-of-thought data, enabling models to mimic human reasoning processes. However, these methods present significant challenges for current general-purpose MLLMs, as they require the model to develop reasoning skills while retaining prior capabilities, which often results in only modest performance gains. Additionally, the lack of structured, high-quality training data impedes training models with advanced reasoning capabilities.

To fully leverage the reasoning capabilities of MLLMs, we propose Insight-V, a novel system comprising two MLLMs dedicated to reasoning and summarization, respectively. The \texttt{reasoning model} is tasked with generating a detailed reasoning process to assist in problem-solving, while the \texttt{summary model} evaluates this reasoning as supplementary information to assess its relevance and utility for answering the question. We also construct a structured, high-quality dataset to train both agents. We posit that this multi-agent system can enhance the reasoning strengths of MLLMs by decomposing the problem-solving process into distinct reasoning and summarization phases, thereby driving substantial performance improvements.

\begin{figure}[t]
\centering
\includegraphics[width=0.5\textwidth]{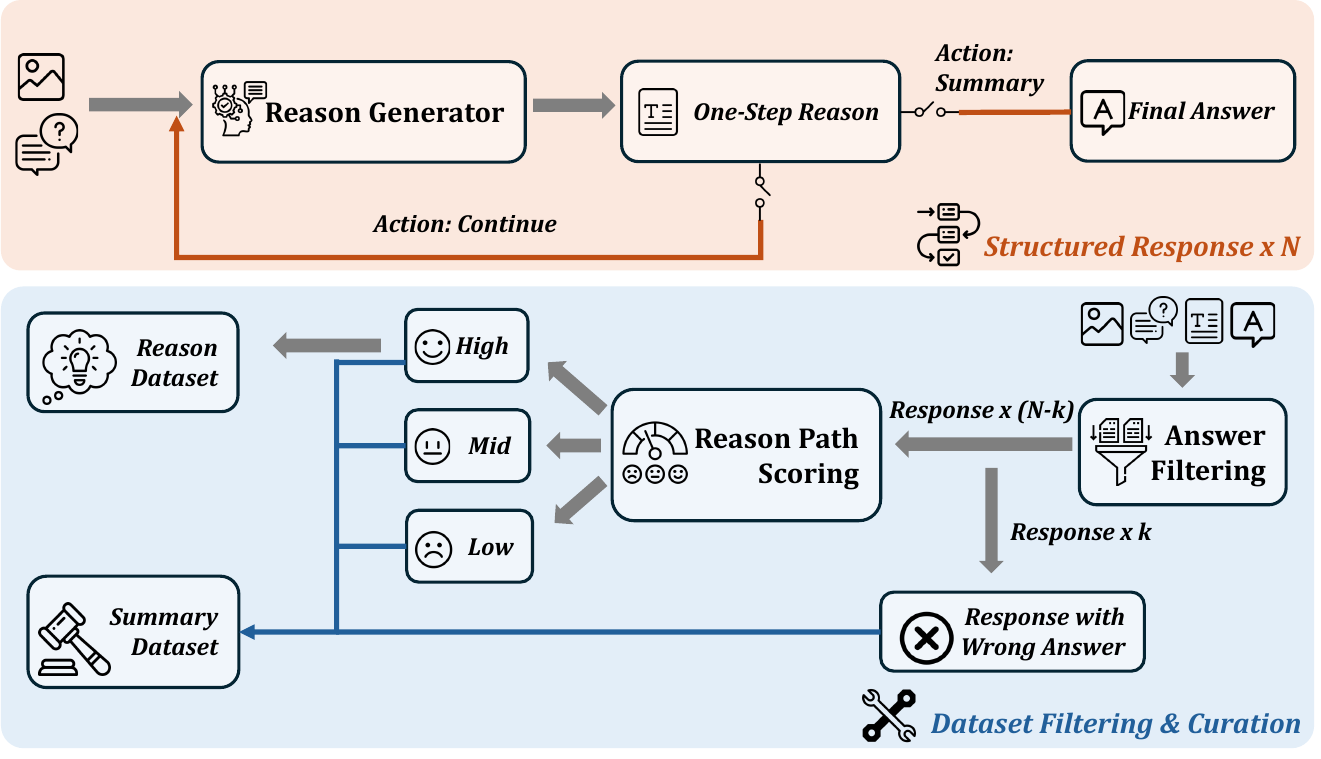} 
\caption{\textbf{Data Generation Pipeline of Insight-V. } The reasoning processes are generated progressively through a reasoning generator, and then fed into a multi-granularity assessment system to ensure high-quality reasoning.}
\label{fig:pipeline}
\vspace{-10pt}
\end{figure}

\subsection{Construction of Structured Reasoning Data}
\label{sec:data}
Previous studies~\citep{zhang2023multimodal,zhang2024mavis} have explored the integration of reasoning capabilities into MLLMs. However, training MLLMs to develop robust reasoning skills remains a considerable challenge, particularly due to data limitations. To address this, we introduce our data generation pipeline in this section, designed to produce high-quality, long-chain reasoning data using a progressive generation process and multi-granularity assessment. As shown in Figure~\ref{fig:pipeline}, this scalable approach enables us to generate high-quality data to enhance the model's reasoning capabilities effectively.

\paragraph{Progressive Long-Chain Reasoning Data Generation. } For each input query, we first employ a reasoning generator to produce a structured reasoning process in JSON format to address the problem. At each step, the reasoning generator provides a brief summary of the current step, a detailed reasoning response, and an action for the following step. If the action is \( continue \), the model proceeds with an additional reasoning step in the next iteration; if the action is \( summary \), the model generates a final summary and answer based on the complete reasoning process in the subsequent iteration. Specifically, For a multi-modal model \( M \), input image \( I \), and question \( Q \), the data generation process for each step is represented as follows:
\begin{align*}
&R_{t}  =   M(I, Q, [R_{1} \cdots R_{t-1}], A), \\
&R_{ans}  = M(I, Q, [R_{1} \cdots R_{n}]),
\end{align*}
where \( R_{t} \) and \( R_{ans} \) denote the response at the \( t \)-th step and the final answer, respectively, \( R_{i} \) represents the reasoning generated by the model at the \( i \)-th step, \(n\) represents the total reasoning steps, and \( A \) is the action determined in the previous step. By repeating this process \( N \) times, we can iteratively sample \( N \) structured responses for each query. The generation parameters are adjusted to encourage the model to produce outputs with various information and steps, allowing us to identify the most effective reasoning chain for each question.

\paragraph{Multi-Granularity Assessment. } After obtaining the structured responses, we utilize an assessment pipeline to ensure data quality. Specifically, we first apply a strong LLM, such as Qwen2~\citep{qwen2}, for direct \textbf{answer filtering}. As shown in Figure~\ref{fig:pipeline}, the model is provided with the generated final answer and the ground truth answer and is tasked with determining whether the generated answer is correct, serving as an approximate indicator of the validity of the associated reasoning chain. Once responses with incorrect answers are filtered out, the remaining reasoning processes are passed to a \textbf{reasoning path scoring} agent. Here, an advanced multi-modal model, such as Qwen2-VL~\citep{qwen2vl}, is supplied with the image, question, reasoning path, and ground truth answer, and is prompted to evaluate the reasoning path. The scoring agent assesses each response based on the step-by-step accuracy of the reasoning path and the level of detail in the reasoning. To ensure consistency in scoring across different data samples, we aggregate all responses for each question and process them in a single pass. The model then generates scores for each response, ranging from 1 to 100.

Through the above two steps, we construct a structured, high-quality dataset that provides detailed reasoning for each question, effectively supporting the training of our models.

\subsection{Model Design}
\label{sec:model}
After constructing the dataset, we develop a multi-agent framework to enhance overall reasoning capabilities through collaborative agent interaction. Specifically, we first train a reasoning agent to generate a detailed reasoning process for each problem. Then, a summary agent is employed to answer the question, selectively utilizing the reasoning process based on its assessment. Together, these two agents collaborate to improve reasoning performance effectively.

\vspace{-10pt}
\paragraph{Reasoning Agent. } Previous approaches typically combine reasoning and question-answering within a single process, which poses challenges for MLLMs. Generating a long-chain reasoning process can introduce errors, and directly answering questions based on flawed reasoning often leads to poorer results. To address this, we propose a specialized reasoning agent designed to generate a detailed, step-by-step reasoning process in response to an input query. We construct the reasoning dataset by selecting the highest-scoring reasoning path for each question. After training on this dataset, the model transforms into a reasoning agent with enhanced reasoning capabilities, enabling it to generate more detailed, structured reasoning processes.

\vspace{-10pt}
\paragraph{Summary Agent. }
Summarization plays a critical role in enabling models to accurately answer questions. After generating multi-step reasoning, summarization provides a cohesive understanding of the reasoning process, ultimately guiding the model to the final answer. However, since the response generated by the reasoning agent may contain errors, we develop a summarization model robust to inaccuracies in the reasoning path, selectively incorporating or disregarding elements as needed. This approach maximizes the reasoning model’s effectiveness while minimizing the risk of introducing misleading information.

To enhance the robustness of the summary agent, we carefully curate its training dataset. We utilize the collected dataset, which is comprised of two types: data with optimal reasoning processes and data with flawed reasoning processes for the summarization task. This method prevents the model from simply copying reasoning outcomes and encourages critical evaluation of reasoning quality. To further promote critical analysis by the summary agent, we select flawed reasoning samples based on their performance scores. Specifically, we draw flawed responses from varying score ranges to create a dataset with different levels of error, prompting the model to assess reasoning processes at various granularities. To better align the summary model with the reasoning agent, we also incorporate question-reasoning pairs generated by the reasoning agent to enhance collaboration between the two agents. Additionally, to preserve the original multi-modal capabilities, we supplement the dataset with standard question-answering data to sustain the summary agent's performance in direct question-answering.

\begin{figure}[t]
\centering
\includegraphics[width=0.5\textwidth]{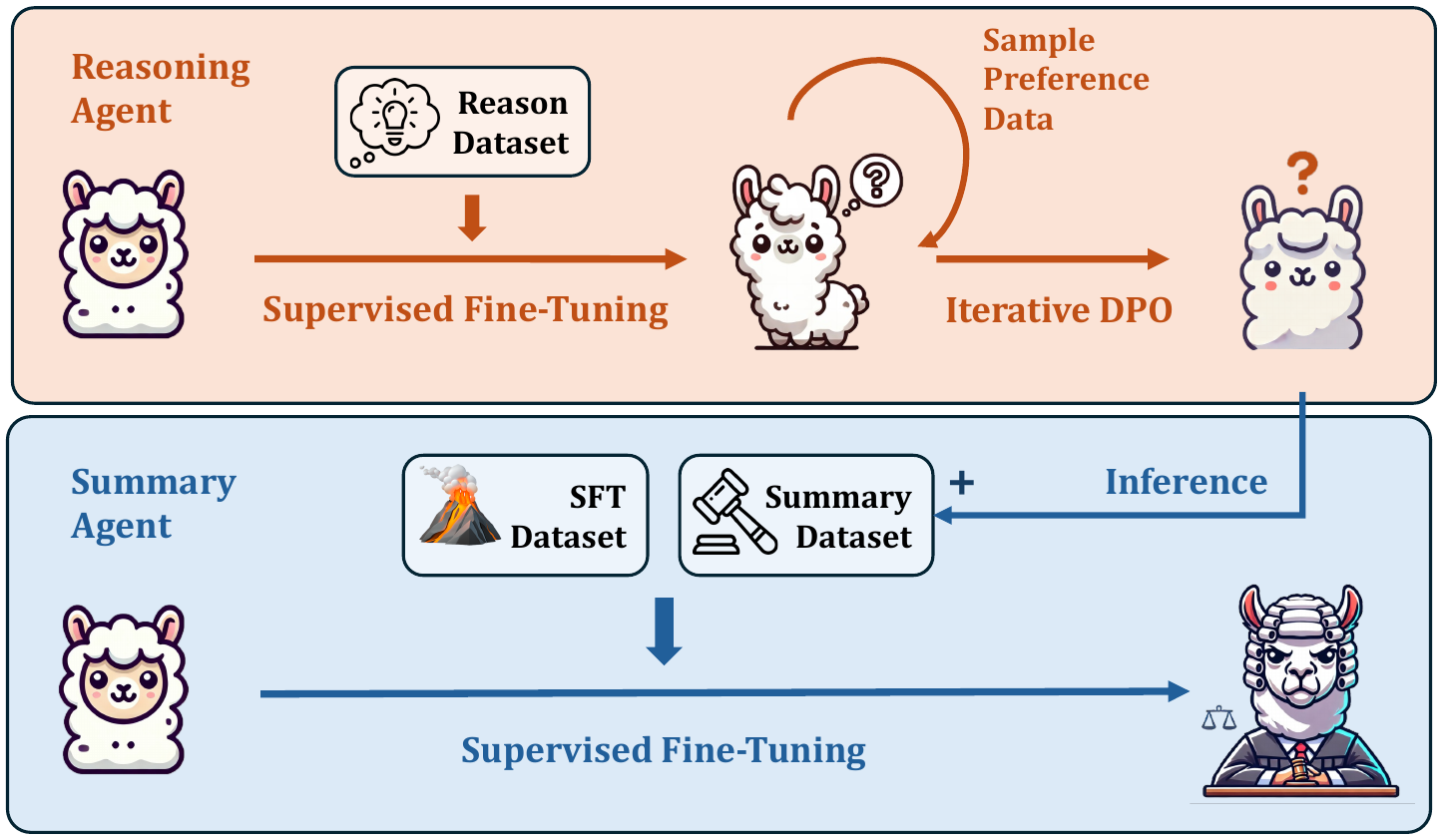} 
\caption{\textbf{Overview of Insight-V Model Design.} We derive a multi-agent system from a single model. By decomposing the task into reasoning and summarization, the two agents collaborate to enhance the overall reasoning capability.}
\label{fig:method}
\vspace{-10pt}
\end{figure}

\subsection{Training Pipeline}
\label{sec:implementation}

The training pipeline for Insight-V is designed to be straightforward and efficient, utilizing a two-stage strategy. For both the reasoning agent and the summary agent, we begin with a well-trained MLLM. In the first stage, we apply supervised fine-tuning to train the agents to fulfill their designated roles using corresponding datasets. In the second stage, we implement direct preference optimization, following prior research~\citep{zhang2024improve,xiong2024llavaovchat}. This optimization is applied to the reasoning model, aligning it with human reasoning processes in a simple yet effective manner. These two stages enable the development of a robust system with enhanced visual reasoning capabilities.

\subsubsection{Supervised Fine-tuning for Multi-agent System}
To perform supervised fine-tuning and obtain the two agents, we first train a base multi-modal model, following established methodologies. This model can address general visual question-answering tasks and gain foundational vision-language skills. We compile a high-quality image-text dataset focused on knowledge learning to train the base model. This data is sourced from various open-source academic datasets, including LLaVA-NeXT~\citep{liu2024llavanext}, Cauldron~\citep{laurenccon2024matters}, and Cambrian-1~\citep{tong2024cambrian}. Once the base model is trained, we further fine-tune the two agents, initializing them from the base model. For the reasoning agent, we utilize the curated reasoning dataset to develop step-by-step reasoning capabilities. For the summary agent, we formulate a dataset as outlined in Section~\ref{sec:model} and sample about one million general image-text pairs from the dataset used for the base model, preserving its original visual perception abilities.

\subsubsection{Enhanced Reasoning with RL}

Preference learning has gained increasing focus in the field of large language models. The primary aim is to fine-tune model outputs to align better with human (or expert) preferences, creating outputs more suited to real-world applications. Let’s assume a preference dataset defined as \(\mathcal{D} = \{(x^{(i)}, y_w^{(i)}, y_l^{(i)})\}_{i=1,\dots,|\mathcal{D}|}\), where each \(x^{(i)}\) is prompt, and \(y_w^{(i)}\) and \(y_l^{(i)}\) represent the preferred and less preferred responses, respectively. We denote \(y_w \succ y_l \mid x\) to signify that \(y_w\) is preferred over \(y_l\) for prompt \(x\).

Since the true distribution of human preferences cannot be directly observed, we approximate it with a latent reward model \(r^*(x, y)\), assuming that higher rewards correspond to stronger preferences. Following the approach by~\citet{rafailov2024direct}, we can model the human preference distribution \(p^*\) using the Bradley-Terry (BT) model~\citep{bradley1952rank}:
\[
\begin{aligned}
p^*(y_1 \succ y_2 \mid x) &= \frac{\exp(r^*(x, y_1))}{\exp(r^*(x, y_1)) + \exp(r^*(x, y_2))} \\
&= \sigma(r^*(x, y_1) - r^*(x, y_2)),
\end{aligned}
\]
where \(\sigma\) denotes the logistic function. 

To estimate the parameters of the reward model, we can apply maximum likelihood estimation by minimizing the negative log-likelihood:
\[
\mathcal{L}_R(r_\phi, \mathcal{D}) = - \mathbb{E}_{(x, y_w, y_l) \sim \mathcal{D}} [\log \sigma(r_\phi(x, y_w) - r_\phi(x, y_l))],
\]
where \(r_\phi\) is a parameterized reward model. This approach allows us to approximate the preference distribution and fine-tune the model to capture human-like preferences effectively.

The traditional DPO algorithm operates in an offline setting. During DPO training, as model parameters continuously change, the preference dataset generated offline can gradually diverge from the model’s current distribution, which weakens the effectiveness of the DPO algorithm. To address this issue, we employ an iterative DPO algorithm. By conducting multiple rounds of DPO training and sampling, this approach enables the model to better approximate an online setting during training, thus further enhancing its performance. Specifically, our approach involves training a sequence of models \( M_1, \ldots, M_T \), where each subsequent model \( M_{t+1} \) utilizes preference data \( \mathcal{D}_t \) generated by the \( t \)-th model. We apply this complete training process to the fine-tuned reasoning agent, enabling the model to better align with human preferences and produce structured, detailed reasoning steps for complex questions, which supports the summary agent more effectively.
\section{Experiments}
\label{sec:exp}

\begin{table*}[t]
  \centering
  \caption{\textbf{Results on Visual Reasoning Tasks.} We conduct evaluation experiments across 7 benchmarks, covering both general reasoning and task-specific reasoning assessments. Insight-V exhibits notable effectiveness and generalizability when applied to LLaVA-NeXT and our baseline model, surpassing other state-of-the-art MLLMs by a large margin.}
    \resizebox{\linewidth}{!}{
    \begin{tabular}{L{150pt}C{30pt}C{45pt}C{60pt}C{40pt}C{55pt}C{45pt}C{45pt}C{45pt}C{50pt}}
    \toprule
    Model & Size & MMMU & MMMU-Pro & MMBench & MME & ChartQA & MMStar & MathVista  &  \cellcolor{blue!10}Average \\
    \midrule 
    DeepSeek-VL~\citep{lu2024deepseek} & 7B & 35.4 & - & 73.5 & -/- & 59.1 & 37.1 & 36.1 & - \\
     VILA-1.5~\citep{lin2023vila} & 8B & 38.6 & - & 75.3 & 1634.9/- & - & 39.7 & - & - \\
     Cambrian-1~\citep{tong2024cambrian} & 8B & 42.7 & - & 75.9 & 1547.1/- & 73.3 & - & 49.0  & -\\
    InternLM-XComposer2~\citep{dong2024internlm} & 7B & 41.1 & - & 77.6 & 2220.4 & 71.8 & 56.2 & 59.5 & - \\
    POINTS~\citep{liu2024points} & 7B & 51.4 & - & 78.0 & 2184.1 & - & 60.9 & 63.0 & - \\
    IXC-2.5~\citep{internlmxcomposer2_5} & 7B & 42.9 & - & 79.4 & 2233.1 & 82.2 & 59.9 & \textbf{63.7} & - \\
    Bunny-LLaMA3~\citep{he2024bunny} & 8B & 43.4 & - & 77.2 & 1588.9/321.1 & - & - & 34.4 & - \\
    MM-1.5~\citep{zhang2024mm1} & 7B & 41.8 & - & - & 1514.9/346.4 & 78.6 & - & 47.6 & - \\
    MiniCPM-LLaMA3-V 2.5~\citep{yao2024minicpmv} & 8B & 45.8  & 19.6 & 77.2 & 2024.6 & - & 51.8 & 54.3 & -\\
        MiniCPM-V-2.6~\citep{yao2024minicpm} & 7B & 49.8 & \textbf{27.2} & 78.0 & 2268.7 & - & 57.5 & 60.6 & - \\
        Qwen2-VL~\citep{qwen2vl} & 7B & \textbf{53.7} & - & 81.0 & - & \textbf{83.0} & 60.7 & 61.4 & - \\
    Idefics3-LLaMA3~\citep{laurenccon2024building} & 8B & 46.6 & 22.9 & 77.5 & 1937.4 & 74.8 & 55.9 & 58.4 & 48.1\\
    Ovis1.5-LLaMA3~\citep{lu2024ovis} & 8B & 48.3 & 23.6 & 76.6 & 1948.5 & 76.4 & 57.3 & 63.0 & 49.4 \\
    \midrule
    LLaVA-NeXT-LLaMA3~\citep{liu2024llavanext} & 8B & 36.9 & 13.2 & 72.3 & 1611.1/346.0 & 69.4 & 43.1 & 45.9 & 40.2\\
    + Multi-Agent & 8B & 40.8 & 17.8 & 77.6 & 1603.7/469.3 & 74.6 & 52.6 & 47.4 & 44.5\\
    \rowcolor{Gray} + Iterative DPO (\textcolor{Red}{\textbf{Insight-V-LLaVA}}) & 8B & 42.0 & 21.0 & 81.7 & 1583.9/485.4 & 77.4 & 57.4 & 49.8 & \textbf{47.2} \textcolor{Red}{\textbf{(+7.0)}}\\
    \midrule
    Our Base Model & 7B & 47.1 & 22.6 & 81.3 & 1573.7/482.5 & 75.7 & 57.0 & 56.9 & 48.7 \\
    + Multi-Agent & 7B & 49.7 & 23.8 & 82.2 & 1662.2/\textbf{629.3} & 81.2 & 58.6 & 58.7 & 50.7\\
    \rowcolor{Gray} + Iterative DPO (\textcolor{Red}{\textbf{Insight-V}}) & 7B & 50.2 & 24.9 & \textbf{82.3} & \textbf{1685.1}/627.0 & 81.5 & \textbf{61.5} & 59.9 & \textbf{51.6} \textcolor{Red}{\textbf{(+2.9)}}\\
    
    \bottomrule
    \end{tabular}%
    }
  \label{tab:reason} \vspace{-10pt}
\end{table*}%

We conduct extensive experiments across multiple vision-language benchmarks to validate the effectiveness of our method. In this section, we first introduce the implementation details of Insight-V in Section~\ref{sec:detail}. Then we present a comparison with state-of-the-art MLLMs, outlining the primary results of our method on visual reasoning tasks as well as additional results on general image understanding in Section~\ref{sec:exp-visual}. Moreover, we offer further analytical experiments and essential ablation studies on design choices in Section~\ref{sec:analysis}, along with qualitative results shown in Section~\ref{sec:qualitative} for more insights.

\begin{table}[t]
  \centering
  \caption{\textbf{Results on other multimodal benchmarks.} Insight-V enhances reasoning capabilities without compromising general visual perception and even achieves improvements on benchmarks requiring perception ability more.}
    \resizebox{\linewidth}{!}{
    \begin{tabular}{l|cccc}
    \toprule
    Model & TextVQA & DocVQA & OCRBench & AI2D \\
    \midrule
    LLaVA-NeXT-LLaMA3 & 65.2 & 78.2 & 553 & 71.5 \\
    + Multi-Agent & 68.9 & 81.8 & 631 & 75.7 \\
    \rowcolor{Gray} + Iterative DPO (\textcolor{Red}{\textbf{Insight-V-LLaVA}}) & \textbf{70.5} & \textbf{82.9} & \textbf{663} & \textbf{77.3} \\
    \midrule
    Our Base Model & 75.4 & 90.2 & 713 & 79.7 \\
    + Multi-Agent & \textbf{77.0} & 91.4 & \textbf{738} & \textbf{80.1} \\
    \rowcolor{Gray} + Iterative DPO (\textcolor{Red}{\textbf{Insight-V}}) & 76.8 & \textbf{91.5} & 735 & 79.8 \\
    
    \bottomrule
    \end{tabular}%
    }
  \label{tab:general} \vspace{-10pt}
\end{table}%

\subsection{Implementation Details}
\label{sec:detail}
We integrate the Insight-V system with various MLLMs to demonstrate the broad applicability of our approach. Our initial implementation with Insight-V on LLaVA-NeXT-LLaMA3~\citep{liu2024llavanext} illustrates the method's effectiveness. To further validate its generalizability and establish a solid baseline against state-of-the-art MLLMs, we additionally train a base multi-modal model using the Qwen-2.5-7B~\citep{qwen2.5} LLM. During pretraining, we utilize the 558K captioning dataset from LLaVA-1.5~\citep{liu2024llava15}, unfreezing the connector parameters. This is followed by supervised fine-tuning with a curated dataset of approximately 4 million images, using a learning rate of 2e-5 as guided by prior researches~\citep{liu2024llava15,liu2024llavanext}. This two-stage training process equips the baseline model with essential visual perception abilities, achieving competitive results on vision-language benchmarks.

We then initialize two agents from the baseline model, performing targeted fine-tuning to obtain the final agents. For the reasoning agent, we compile a dataset of 200K images and train the model over 2 epochs with a learning rate of 5e-6. For the summary agent, we use a dataset of 1.2 million images, applying a learning rate of 1e-5 and training for 1 epoch. Additionally, we apply Direct Preference Optimization (DPO) to the reasoning agent, using approximately 15K preference data and training for 1 epoch at a learning rate of 5e-7. This DPO process is iteratively conducted across 3 rounds by using the model from the previous stage to generate preference data, thus enhancing the agent's reasoning capabilities. The lmms-eval~\citep{li2024xinrun} is utilized for fast evaluation. Further training details are provided in the appendix.

\subsection{Main Results on Visual Reasoning}
\label{sec:exp-visual}
\paragraph{Setup.} We present comprehensive experimental results on various visual reasoning benchmarks, which require the model to handle complex and challenging questions, demonstrating strong reasoning capabilities. We select several well-known and representative benchmarks, covering comprehensive evaluation, chart understanding, and mathematical problem-solving. MMMU~\citep{yue2024mmmu} and MMMU-Pro~\citep{yue2024mmmupro} evaluate the model’s expert-level perception and reasoning abilities across various topics. MMBench~\citep{liu2023mmbench} is a widely used benchmark for comprehensive MLLM evaluation on images. MME~\citep{fu2023mme} includes 14 challenging subtasks for assessing visual perception and cognition. ChartQA~\citep{masry2022chartqa} focuses on logical reasoning with charts, while MathVista~\citep{lu2023mathvista} assesses problem-solving skills in math, using GPT-4-0613~\citep{OpenAI_GPT4_2023} as the evaluator, following standard practice. MMStar~\citep{chen2024we} tests a range of tasks with varying difficulty levels.

\begin{table}[t]
  \centering
  \caption{\textbf{Ablations on the Insight-V Design Choice.} The multi-agent design outperforms other configurations, highlighting the critical role of reasoning and summarization decomposition.}
    \resizebox{\linewidth}{!}{
    \begin{tabular}{l|ccccc}
    \toprule
    Model & MMMU & ChartQA & MathVista & MMStar & Avg \\
    \midrule
    Baseline & 47.1 & 75.7 & 56.9 & 57.0 & 59.2 \\
    Vanilla - Direct SFT & 47.0 & 79.2 & 57.6 & 58.4 & 60.6 \\
    Multi-Turn Supervised & 48.1 & 79.6 & 57.9 & 58.2 & 61.0 \\
    Summary Agent Only & 47.5 & 76.3 & 57.3 & 57.9 & 59.8 \\
    \rowcolor{Gray} Multi-Agent & 49.7 & 81.2 & 58.7 & 58.6 & 62.1 \\
    
    \bottomrule
    \end{tabular}%
    }
  \label{analysis:design} \vspace{-10pt}
\end{table}%

\paragraph{Main Results.} The experimental results in Table~\ref{tab:reason} confirm the effectiveness and generalizability of Insight-V on visual reasoning tasks. Applied to LLaVA-NeXT and our baseline model, the Insight-V system substantially boosts performance across challenging visual reasoning benchmarks, achieving competitive outcomes against state-of-the-art methods. We report an average improvement across all benchmarks: Insight-V yields an average increase of 4.9\% and 3.2\% when applied to LLaVA-NeXT and the base model, underscoring the system’s effectiveness. Following Direct Preference Optimization, an additional improvement of 2.6\% and 1.0\% was observed, illustrating the value of reinforcement learning algorithms in enhancing reasoning capabilities. Specifically, on general reasoning benchmarks, MMMU, MMMU-Pro, and MMBench, we noted gains of 3.1\%, 2.3\%, and 1.0\%, respectively. The highest improvement was on MME, with a 9.1\% increase, highlighting Insight-V’s strength in tasks requiring both perception and reasoning. On ChartQA, which entails complex chart-based reasoning, we achieved a 5.8\% increase. Performance gains on MMStar and MathVista, with improvements of 4.5\% and 3.0\%, further validate our approach’s versatility in tackling math problems that demand sophisticated reasoning and numerical calculation skills. This strong performance across all benchmarks underscores the efficacy of Insight-V and its potential to advance visual reasoning capabilities.


\paragraph{Results on Other Multi-modal Benchmarks.}  To verify that Insight-V enhances reasoning capabilities without compromising basic visual perception skills, we conduct experiments on benchmarks requiring only fundamental image understanding. We selected TextVQ~\citep{singh2019textvqa}, DocVQA~\citep{mathew2021docvqa}, OCRBench~\citep{liu2023ocrbench}, and AI2D~\citep{kembhavi2016ai2d} for this purpose, as these benchmarks focus on basic visual interpretation rather than complex reasoning.

As shown in Table~\ref{tab:general}, integrating Insight-V enables models to maintain strong performance across general visual perception benchmarks. Additionally, applying Insight-V yields performance improvements on both LLaVA-NeXT and our base model, as it enhances general visual understanding by guiding the model’s attention to relevant regions, thereby improving visual perception accuracy. Following DPO, the model demonstrates advanced reasoning capabilities while achieving competitive results on standard image understanding benchmarks compared to baseline models, underscoring Insight-V's robustness. Consistent results of both models further confirm Insight-V’s generalizability.

\begin{figure}[!t]
\centering
\includegraphics[width=\linewidth]{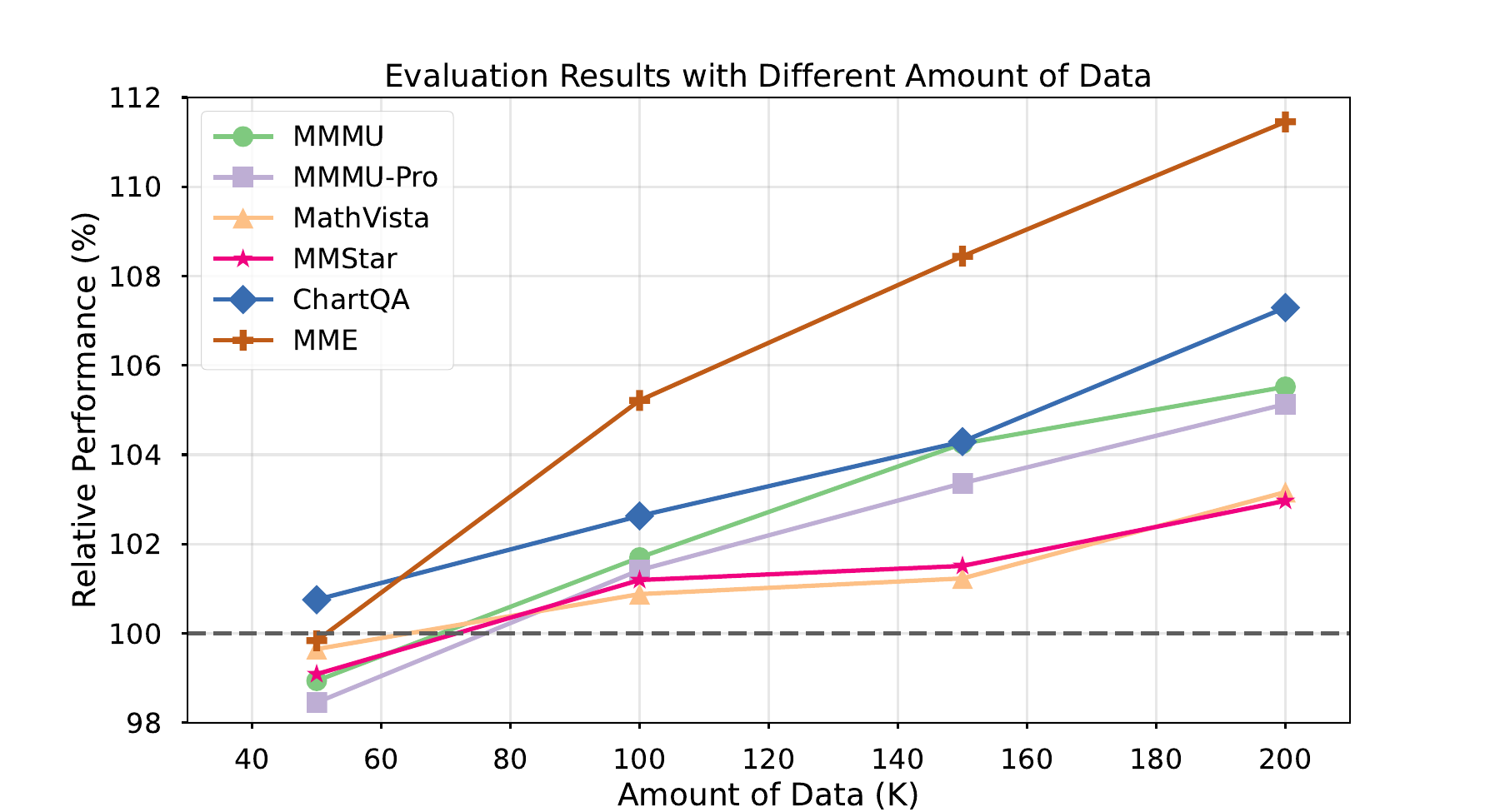} 
\caption{\textbf{Ablations on the amount of training data.} The reasoning agent benefits from data scaling, providing more valuable insights for the summary agent.}
\label{fig:scaling}
\vspace{-10pt}
\end{figure}

\subsection{Further Analysis}
\label{sec:analysis}
In this section, we present comprehensive experiments to validate the design choices of Insight-V, emphasizing our approach's key contributions. Additionally, we include a case study to further demonstrate the qualitative effectiveness of Insight-V.

\paragraph{Effectiveness of Multi-agent System.} To evaluate the effectiveness of the multi-agent system, we comprehensively compare Insight-V with alternative design choices. We begin by reporting the performance of the summary agent without a reasoning process to underscore the importance of the reasoning agent. For a comprehensive comparison, we also restructure the collected data using a Chain-of-Thought (CoT) template and train a model to perform sequential reasoning and subsequently answer questions. Additionally, we enable a model that can handle multi-turn conversations, facilitating reasoning and summarization in a multi-turn format. We use MMMU, ChartQA, MathVista, and MMStar as representative benchmarks to assess the performance across different methods.

\begin{table}[t]
  \centering
  \caption{\textbf{Ablations on the DPO training strategy.} Iterative DPO progressively enhances the model's reasoning capabilities, leading to improved performance.}
    \resizebox{\linewidth}{!}{
    \begin{tabular}{l|ccccc}
    \toprule
    Model & MMMU & ChartQA & MathVista & MMStar & Avg \\
    \midrule
    Insight-V (Multi-Agent) & 49.7 & 81.2 & 58.7 & 58.6 & 62.1 \\
    + RLAIF & 49.5 & 81.4 & 59.1 & 59.2 & 62.3 \\
    + DPO & 50.8 & 80.8 & 59.3 & 59.9 & 62.7 \\
    \rowcolor{Gray} + Iterative DPO & 50.2 & 81.5 & 59.9 & 61.5 & 63.3 \\
    \bottomrule
    \end{tabular}%
    }
  \label{analysis:dpo} \vspace{-10pt}
\end{table}%

As shown in Table~\ref{analysis:design}, the results indicate that the multi-agent system plays a critical role in enhancing the system’s visual understanding capabilities. Using only the summary agent without a reasoning process results in limited improvements in reasoning tasks, as the model lacks the necessary reasoning framework for optimal performance. Training the model to perform chain-of-thoughts reasoning, which is denoted as Insight-V (Vanilla-Direct SFT), yields modest gains, as this approach does not sufficiently emphasize critical judgments within the reasoning process, merging reasoning and summarization into a single task. Furthermore, models trained with multi-turn conversations still produce sub-optimal results, underscoring the importance of a multi-agent system that separates reasoning and summarization.

\begin{figure*}[t]
\centering
\includegraphics[width=0.98\textwidth]{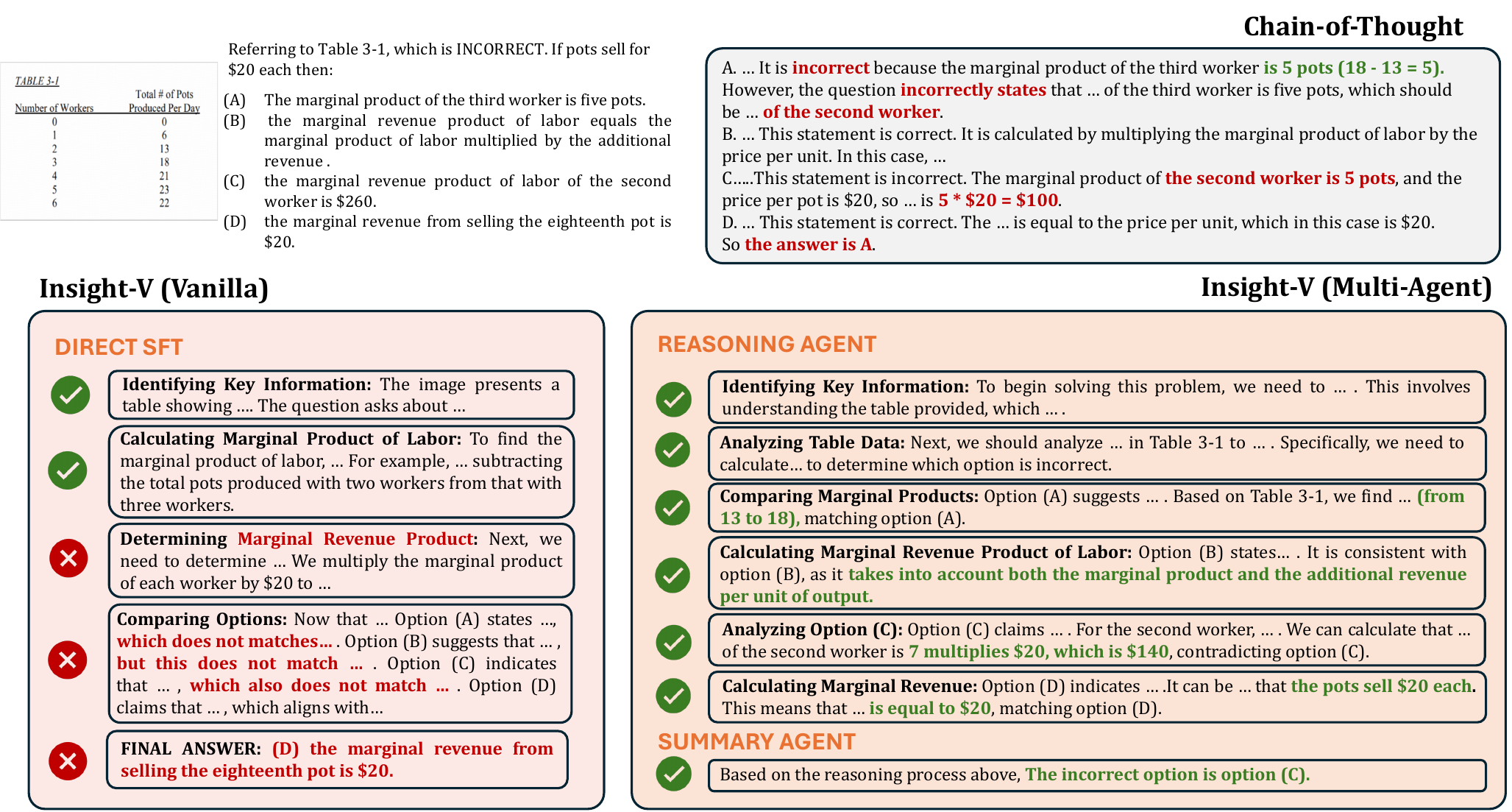}
\vspace{-5pt}
\caption{\textbf{Qualitative Results of Insight-V.} We present qualitative comparisons of Insight-V with Chain-of-Thought and learning  Insight-V with direct SFT (Vanilla). For the Insight-V system, the reasoning agent delivers a more coherent and structured reasoning process, guiding the summary agent toward the correct answer, whereas other methods struggle with complex reasoning tasks and fail to solve such challenging problems. }
\vspace{-10pt}
\label{fig:showcase}
\end{figure*}

\paragraph{Data Scaling Law of Reasoning Agent.} To assess the effectiveness of the reasoning agent, we perform ablation experiments on the data volume used for training. As shown in Figure~\ref{fig:scaling}, we compare reasoning agents trained on varying data sizes: 50K, 100K, 150K, and 200K samples. For a fair and comprehensive comparison, we employ the same summary agent across evaluations and report results on six benchmarks. The findings clearly indicate that the reasoning agent benefits from increased data. With limited data, the reasoning agent struggles to generalize and fails to provide useful input to the summary model, resulting in performance even worse than baseline models. Conversely, training on larger datasets enhances the reasoning agent’s capabilities, enabling it to perform step-by-step reasoning and provide valuable insights that support the summary agent in solving tasks effectively.

\paragraph{Effects of RL Algorithms. } We assess DPO algorithms to identify optimal alignment strategies for enhancing the reasoning process. To examine the effects of dataset composition on DPO training, we compare our curated dataset against the widely-used RLAIF-V~\citep{yu2024rlaif} dataset, which contains 80K DPO data pairs for alignment. We also investigate the potential benefits of iterative DPO in advancing the model's reasoning capabilities. For an unbiased comparison, we subsample the RLAIF-V dataset to approximately 15K preference data points, aligning it with the size of our dataset. The results, summarized in Table~\ref{analysis:dpo}, show an average improvement of approximately 0.2\% on the evaluated benchmarks when trained with the RLAIF-V dataset, whereas our curated dataset achieves greater gains of 0.6\%. This suggests that a DPO dataset based on model-generated rationales, rather than externally sourced data, more effectively boosts reasoning accuracy. Moreover, conducting two additional rounds of DPO training using the same methodology yields further performance gains of 0.6\%, underscoring that iterative DPO training progressively refines the model's ability to generate precise and high-quality reasoning processes compared to a single training pass.

\subsection{Qualitative Results}
\label{sec:qualitative}

We present qualitative comparisons in Figure~\ref{fig:showcase} to illustrate the improvements introduced by integrating the Insight-V system. Specifically, we provide an example that highlights advancements in the reasoning process. In this example, the model is challenged with a multiple-choice question based on a table. We compare our approach with a direct Chain-of-Thought application, and the model undergoes supervised fine-tuning without incorporating a multi-agent system, which is denoted as Insight-V (Vanilla). It is evident that applying Chain-of-Thought directly results in suboptimal reasoning and leads to incorrect answers. Although fine-tuning a single model yields somewhat better reasoning, the model begins to falter as the reasoning chain lengthens, ultimately arriving at incorrect answers. This outcome arises because the model is required to generate both reasoning steps and the final answer simultaneously, limiting its judgment capabilities and weakening its robustness in handling flawed reasoning chains.  In contrast, employing Insight-V enables more logical step decomposition and a structured reasoning chain. The model can analyze each option step-by-step and perform detailed calculations, which the other two methods cannot achieve. The fine-tuned summary model is able to evaluate the reasoning process and determine whether to derive the final answer based on it, significantly enhancing system robustness and ensuring correct answers.
\section{Conclusion}
\label{sec:conclusion}
In this paper, we introduce Insight-V, a novel system that combines a scalable data generation pipeline for long-chain, high-quality reasoning data and an effective multi-agent training pipeline to enhance the reasoning capabilities of MLLMs. By developing this system, we provide a scalable approach to model training aimed at improving reasoning performance. Our extensive evaluation across various benchmarks demonstrates the effectiveness of our approach, paving the way for equipping MLLMs with enhanced reasoning capabilities.

\section*{Acknowledgement}
This study is supported by the Ministry of Education, Singapore, under its MOE AcRF Tier 2 (MOE-T2EP20221-0012, MOE-T2EP20223-0002), and under the RIE2020 Industry Alignment Fund – Industry Collaboration Projects (IAF-ICP) Funding Initiative, as well as cash and in-kind contributions from the industry partner(s).

\appendix
\section*{Appendix}

\section{More Implementation Details}
\label{supp:implementation}
In this section, we provide a detailed explanation of the implementation of the DPO strategy. To collect preference data, we sample 16 outputs for each image-text pair to ensure diversity and maintain data quality. Each question, along with its ground truth answer and corresponding reasoning processes, is then presented to advanced LLMs such as Qwen2.5-72B. The model evaluates all reasoning paths in a single forward pass, assigning scores to each. Reasoning paths with scores above 85 are selected as positive examples. To increase the task's complexity, we do not use the lowest-scoring reasoning path as the rejected example. Instead, we choose a reasoning path with a score around 25, ensuring that the DPO-trained model does not overfit specific data patterns. During DPO training, the parameter $\beta$ is set to 0.1, and a standard supervised fine-tuning loss is incorporated to stabilize the training process.

\section{Analysis Experiments of Multi-agent System}
\label{supp:analysis}
To further validate the effectiveness of the proposed multi-agent system, we conduct additional analysis experiments highlighting the superior performance of Insight-V.
\paragraph{Insight-V Generates More Accurate Reasoning Paths and Demonstrates Robustness to Flawed Reasoning.} To demonstrate that the summary model of Insight-V effectively evaluates the quality of reasoning paths and selectively answers questions based on these paths, we conduct analysis experiments on MMStar. These experiments highlight why Insight-V benefits from the integration of the summary agent, leading to improvements in performance.

\begin{figure}[h]
\centering
\includegraphics[width=\linewidth]{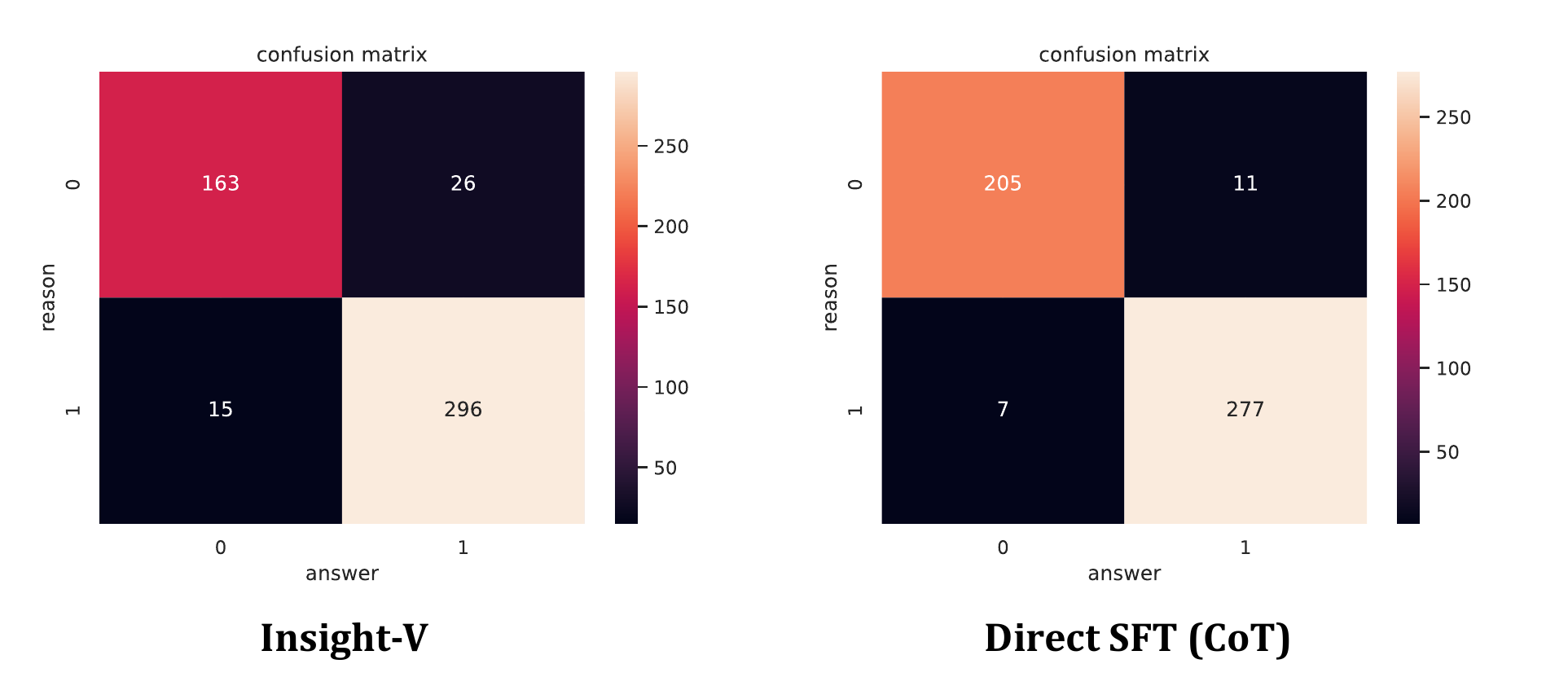} 
\caption{\textbf{Analysis of Multi-agent System.} Insight-V enhances reasoning capabilities while enabling the ability to selectively answer questions based on the provided reasoning process.}
\label{fig:analysis}
\vspace{-10pt}
\end{figure}

As illustrated in Figure~\ref{fig:analysis}, we compute the confusion matrix for the reasoning path and the final answer. A reasoning path is classified as \texttt{True Positive} if both the reasoning path and the final answer are correct, represented in the bottom-right corner of the matrix. Conversely, if the reasoning path is incorrect but the final answer is correct, it is categorized as \texttt{False Negative}, shown in the upper-left corner of the matrix. The results clearly demonstrate that Insight-V generates more accurate reasoning paths, as depicted in the figure. Moreover, even when the reasoning path is incorrect, Insight-V is still capable of producing the correct final answer, showcasing its superior ability to selectively utilize reasoning paths compared to direct fine-tuning with Chain-of-Thought data.

\section{Discussion and Limitations}
\label{supp:discuss}
Insight-V represents an initial exploration into building models capable of o1-like reasoning. Our findings indicate that leveraging MLLMs to perform single-step reasoning and organizing these steps into structured, long-chain reasoning paths, is a promising approach. After fine-tuning on this dataset, the model demonstrates the ability to perform long-chain reasoning. Additionally, we implement a multi-agent system to decompose the question-answering process into distinct reasoning and summarization stages, enabling the system to focus on reasoning while selectively incorporating its results into the summarization process.

As an early attempt to develop robust reasoning models, we acknowledge several limitations that require future improvements to create systems capable of matching the performance of GPT-o1. First, enhancing sampling efficiency is critical. Currently, the process depends on other models for multi-granularity assessment, which could be made more efficient by evaluating reasoning results at each step and pruning redundant samples. This would streamline the system and improve its overall efficiency.

Furthermore, training two models of the same size may not be scalable. Improving the reasoning agent could allow for training a smaller, cost-effective summarization agent, as summarization is inherently a less complex task than reasoning. This adjustment would not only reduce resource requirements but also improve the system's scalability.

Improving overall reasoning quality is equally critical. This includes enabling the model to reflect on previous reasoning steps or implementing inference scaling strategies, which can provide a stronger foundation for the summary model to effectively answer questions.

In conclusion, we hope our method serves as a foundational attempt to inspire and guide future research in this emerging and exciting field.

{
    \small
    \bibliographystyle{ieeenat_fullname}
    \bibliography{main}
}


\end{document}